\RequirePackage{amsmath}
\documentclass[runningheads]{llncs}
\usepackage[T1]{fontenc}
\usepackage{amsmath}
\usepackage{hyperref}
\usepackage{algorithm}
\usepackage{algpseudocode}
\usepackage{booktabs}
\usepackage{graphicx}
\usepackage{listings}
\usepackage{xcolor}

\definecolor{codegreen}{rgb}{0,0.6,0}
\definecolor{codegray}{rgb}{0.5,0.5,0.5}
\definecolor{codepurple}{rgb}{0.58,0,0.82}
\definecolor{backcolour}{rgb}{0.95,0.95,0.92}

\lstdefinestyle{mystyle}{
    backgroundcolor=\color{backcolour},   
    commentstyle=\color{codegreen},
    keywordstyle=\color{magenta},
    numberstyle=\tiny\color{codegray},
    stringstyle=\color{codepurple},
    basicstyle=\ttfamily\footnotesize,
    breakatwhitespace=false,         
    breaklines=true,                 
    captionpos=b,                    
    keepspaces=true,                 
    numbers=left,                    
    numbersep=5pt,                  
    showspaces=false,                
    showstringspaces=false,
    showtabs=false,                  
    tabsize=2
}

\begin{document}

\title{Distilling Event Sequence Knowledge From Large Language Models}

%
%
\author{Somin Wadhwa\inst{1,2}
\and 
Oktie Hassanzadeh\inst{1}
\and 
Debarun Bhattacharjya\inst{1}
\and \\
Ken Barker\inst{1} 
\and 
Jian Ni\inst{1}
}
\authorrunning{Wadhwa et al.}
%
\institute{IBM Research, USA \and
Northeastern University, USA \\
\email{wadhwa.s@northeastern.edu,\{hassanzadeh, debarunb, kjbarker\}@us.ibm.com}
}


\maketitle              
\begin{abstract}
Event sequence models have been found to be highly effective in the analysis and prediction of events. Building such models requires availability of abundant high-quality event sequence data. In certain applications, however, clean structured event sequences are not available, and automated sequence extraction results in data that is too noisy and incomplete. In this work, we explore the use of Large Language Models (LLMs) to generate event sequences that can effectively be used for probabilistic event model construction. This can be viewed as a mechanism of distilling event sequence knowledge from LLMs. Our approach relies on a Knowledge Graph (KG) of event concepts with partial causal relations to guide the generative language model for causal event sequence generation. We show that our approach can generate high-quality event sequences, filling a knowledge gap in the input KG. Furthermore, we explore how the generated sequences can be leveraged to discover useful and more complex structured knowledge from pattern mining and probabilistic event models. We release our sequence generation code and evaluation framework, as well as corpus of event sequence data.
\keywords{Knowledge Graphs, Large Language Models, Knowledge Distillation}
\end{abstract}

\section{Introduction}

Building probabilistic models from event sequence data has numerous practical applications across different domains when plentiful high-quality event data is available. For example, in Finance, event models can be used to predict stock market trends and make informed investment decisions. In Healthcare, event models can help identify patterns and correlations in patient data to improve diagnoses and treatment plans. In the field of Cybersecurity, these models can be used to detect and prevent potential cyber attacks by analyzing the sequence of events leading up to a breach. A common characteristic of the data in these domains is that sequences are clearly associated with an entity (e.g., a company, a person, or a device). There are however other domains where such a clean association between events and entities may not be possible. One such application is news event analysis \cite{CekinelK22,du-etal-2022-resin,HassanzadehA0BB22,RadinskyDM12}. While various news sources record and describe newsworthy events, it is often not possible to automatically put together coherent sequences of events, because each event may involve multiple topics and actors, and many correlated and unrelated events may be occurring simultaneously 
or in close proximity

Prior work has addressed this challenge by devising automated mechanisms for extracting narratives \cite{NorambuenaSurvey23,SantanaSurvey23}, topic detection and tracking~\cite{AllanTDTBook12}, and timeline summarization~\cite{gholipour-ghalandari-ifrim-2020-examining}. While these different categories of solutions have been successful in a range of applications, the outcome is inherently noisy and not in the form of structured event sequences useful for the construction of event models. Prior work has shown little success in automatically turning narratives into structured sequences. In this paper, we explore a novel mechanism for creating structured event sequences in such domains, using generative language models.



\begin{figure}[t]
\centering
  \includegraphics[width=\textwidth]{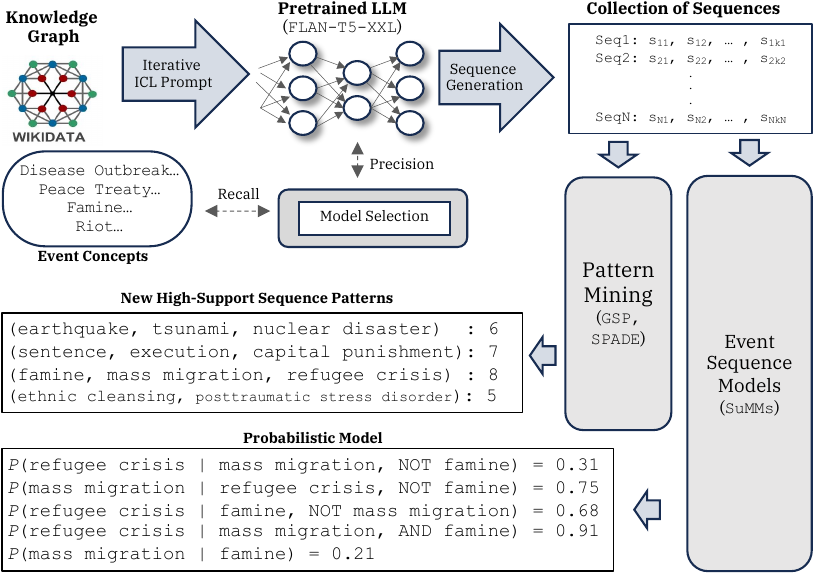}
  \caption{
  An overview of our framework for distilling event sequence knowledge from LLMs, along with examples portraying potential use cases. We show that by \textbf{(1)} starting with a sparse knowledge graph such as Wikidata, we can generate targeted event sequences. Owing to the inherent sparsity in the underlying KG, we can \textbf{(2)} use LLMs to carry out a portion of the evaluation (i.e. precision) to select an optimal model. On this new generated sequence dataset, we then \textbf{(3)} apply classical pattern-mining algorithms (e.g., GSP) to identify potentially interesting \texttt{has$\_$cause} and \texttt{has$\_$effect} event sequence chains, and \textbf{(4)} learn summary Markov models (SuMMs) to identify potential influencing events for particular event types of interest; these are both illustrations of extracting complex structured knowledge from the generated sequences. 
  }
  \label{fig:main}
\end{figure}

Large Language Models \cite{NEURIPS2020_1457c0d6,10.5555/3455716.3455856,wei2022finetuned} have recently become the dominant paradigm in a range of natural language processing (NLP) tasks \cite{chung2022scaling} and often beat traditional approaches on a number of challenging tasks, including complex arithmetic reasoning \cite{imani-etal-2023-mathprompter} and open-domain question answering \cite{kamalloo-etal-2023-evaluating}. In this paper, our goal is to examine the capability of LLMs to generate structured event sequences useful for event analysis. Our aim is to use LLMs directly for generation of event sequences, as opposed to improving prior extraction methods that would rely on high-quality and comprehensive sources describing detailed event sequences for a wide variety of events. Our hypothesis is that LLMs trained on large corpora have already gathered the required knowledge of plausible event sequences and therefore can be suitably guided to produce diverse and high-quality event sequences. To effectively distill this knowledge, we use event-related concepts in Wikidata~\cite{10.1145/2629489}, a comprehensive general-domain knowledge graph, to guide the sequence generation. 
This can be viewed as a novel mechanism for knowledge-guided text generation~\cite{YuKnowledgeEnhancedGenerationSurvey22} and symbolic knowledge distillation~\cite{west-etal-2022-symbolic}. 
We then use these generated sequences for pattern mining and learning probabilistic event models, 
as a way of further structuring the underlying knowledge. Figure~\ref{fig:main} shows our overall framework along with examples from our experiments of patterns mined from an LLM-generated event sequence collection, as a well as a simple model learned from the collection.

In summary, we make the following contributions: 

\begin{enumerate}
    \item We devise a new iterative in-context prompting strategy for generating high-quality event sequences using LLMs. To the best of our knowledge, we are the first to use LLMs to generate structured event sequences for the purpose of analyzing various event models.
    \item We compile high-quality event sequences using our generation mechanism, based on a curated set of high-level event concepts (classes) from Wikidata that represent newsworthy events. 
    \item We develop an evaluation framework and conduct experiments to show the value of LLM-powered event sequence generation on replicating and augmenting knowledge in structured representations such as knowledge graphs.
    \item We further demonstrate the practical usefulness of our approach by leveraging downstream pattern mining and probabilistic event models. 
\end{enumerate}

Our code and generated sequence data are included in the supplementary material, and will be made publicly available for future research.

\section{Related Work}

\paragraph{News Event Analysis} The primary applications of our work are around news event analysis and forecasting. Liang~\cite{LiangEventPredictionSurvey21} presents a taxonomy of different flavors of event prediction in the literature. Our target event prediction applications fall under the ``Semantic Prediction" category, with time and location details not being of interest, and the primary goal being the prediction of ``event profiles" such as event types. Seminal work in this area is the work of Radinsky et al.~\cite{RadinskyDM12} where causal relations between past events are extracted from text and then a knowledge graph is utilized to generalize the extracted relations in order to make predictions. More recent work has explored the use of graph sequence mining over a graph structure representation of events extracted from text~\cite{CekinelK22}, with graph mining used as a mechanism of extracting useful relations from large and noisy outputs of extraction. The application of building event sequence models over such outputs has not been explored. Our paper explores an alternative approach of event sequence generation that is capable of generating longer and potentially higher-quality sequences. 


\paragraph{Event Sequence Extraction from Text}

There is a wealth of literature on different methods of extracting sequences from textual corpora.  Norambuena et al.~\cite{NorambuenaSurvey23} present a comprehensive survey of automated methods of narrative extraction. Narratives contain several elements including events, participants (actors/protagonists), time, and space. The task of event detection itself is a highly challenging task and the topic of extensive research~\cite{chen2021event,xiangSurvey19,li-etal-2020-connecting}, which has its root in the Topic Detection and Tracking (TDT) task \cite{SantanaSurvey23}. This line of work started out as a DARPA-sponsored initiative with the same name~\cite{Allan1998}. Another closely related task is news timeline summarization~\cite{gholipour-ghalandari-ifrim-2020-examining}. While pattern mining algorithms have been applied to the output of such extractions, e.g. for creation of ``domain templates"~\cite{filatova-etal-2006-automatic}, we are not aware of any attempts to use the extractions to construct event models for analysis or prediction. This is mainly due to the fact that the output of such methods often result in very short and noisy sequences, not suitable for the majority of event sequence models.



\paragraph{Sequential Pattern Mining and Event Sequence Models} Sequential pattern mining and related approaches have been the subject of extensive research~\cite{mannila1997,MabroukehSequentialPatternMiningSurvey10,MooneySequentialPatternMiningSurvey13,fournier2017}. 
These algorithms take in a set of sequences (or ``sequential records") with a set of unique events (or ``items") and often a ``minimum support" threshold, and return as output a ranked set of all frequent sequences 
in a given sequence collection (or database) meeting the minimum support threshold. There is also a long line of work on statistical and probabilistic models for analysis of different kinds of event sequences. Our focus in this paper is on multivariate event sequences, i.e., sequences of various event types without timestamps. Markov models for sequences~\cite{raftery1985,BegleiterEY04} and long short-term memory (LSTM)~\cite{HochreiterLSTM97} models are examples of prediction models. We leverage a more recent family of models -- Summary Markov models~\cite{ijcai2022p0670} -- for some of the experiments in this paper that involve analyzing generated event sequences.

\section{Knowledge-Guided Event Sequence Generation}
\label{sec:seq_gen_methods}

Through utilizing LLMs, we model event prediction as a \textit{conditional generation task} under zero and iterative few-shot settings. Our targets are linearized sequences of event concepts. We begin by prompting  a large language model (with in-context exemplars) with an event trigger $y_1$ to generate the next concept from a defined set of labels, and repeat this process until we get a sequence of desired length. Formally, given an event trigger $\mathcal{T}$, we model the probability of generating linearized string $y$ of length $T$ containing $N$ unique event concepts that follow $\mathcal{T}$ in sequence:

\begin{equation}
    p_{\text{LM}}(y | \mathcal{T}) = \prod_{t=2}^{T}p(y_t | \mathcal{T}, y_{<t-1})
\end{equation}

\begin{figure}[ht]
    \centering
\includegraphics[width=3.4in]{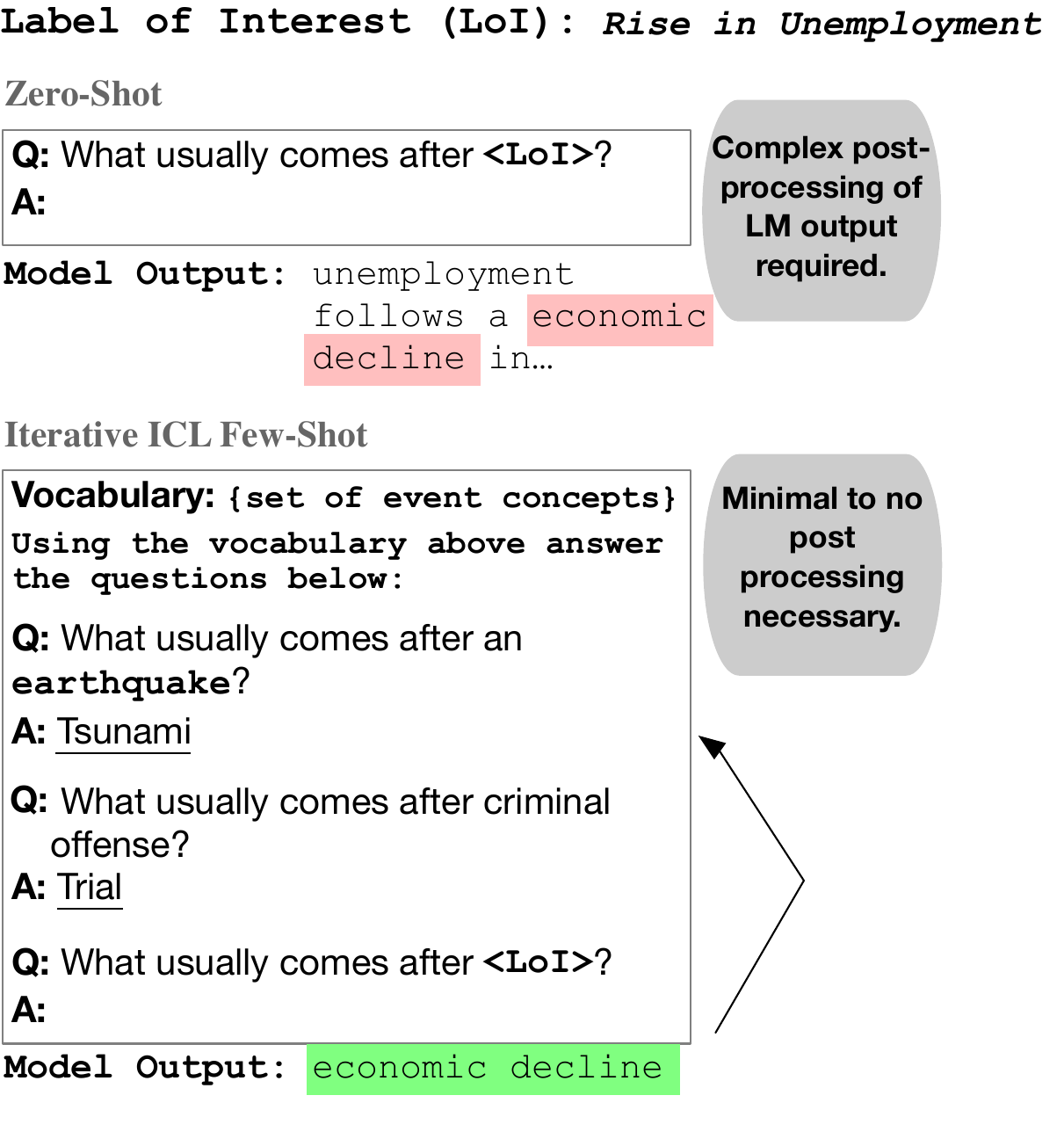}
\caption{Illustration of our approach to elicit event sequence knowledge given a \textit{label of interest}. Use of instructional in-context exemplars substantially reduces the need for post-processing LLM output in addition to constraining the output label space.} 
\label{fig:prompt_examples}
\end{figure}

This is the standard conditional language modeling objective. We try multiple prompting techniques and qualitatively observe optimal results with an \textit{iterative} in-context few-shot prompting strategy (Figure \ref{fig:prompt_examples}). Specifically, we start with a set of six randomly selected examples of the form -- ``What usually follows event $X$?''. This approach follows the incremental prompting procedure from \cite{li-etal-2023-open}. Based on the model output ($Y$) from a pre-specified vocabulary, we append this same example to the original prompt \textit{in conjunction} (i.e. ``What usually follows event $X$ and $Y$?'') with ICL examples of the same form. As shown in Figure \ref{fig:prompt_examples}, our iterative technique serves dual purposes: (i) it leverages in-context learning; and (ii) eliminates the need for implementing complex resolvers to post-process model outputs. We repeat this process until a sequence of a desired minimum length $m$ is achieved (in our experiments, $m=3$) or we've exhausted a maximum number of tries ($k=10$ in our experiments) to generate an in-domain event type.

\paragraph{Identifying Event Concepts} With incremental prompting we curate a new dataset of high-level event concepts (classes) from Wikidata that represent newsworthy events. To do so, we query Wikidata for event concepts  that have links to Wikinews articles and are instances of classes that are a subclass of the \texttt{occurrence} class, i.e., indicating they are newsworthy event classes. We gathered $50$ top-level classes for these event concepts, each having multiple labels (e.g. \texttt{conflict} $\rightarrow$ \texttt {conflict (psychological), dispute, disagreement, etc.}). This yielded a total of $202$ unique event labels for the $50$ top-level classes. Most of these event concepts have \textit{some} causal relations (i.e. \texttt{has\_cause} or \texttt{has\_effect}). We use these relation pairs as in-context exemplars to create our prompts. We then generate event sequences through iterative in-context prompting (Figure \ref{fig:prompt_examples}). Full length prompts used in all our experiments are provided in the Appendix.

To generate new event sequences in a zero-shot setting, we start with an event trigger (e.g. a concept like \texttt{workplace accident}), create a prompt with instructions describing desired relationships and a few in-context exemplars (ICL prompt), and constrained the model output to the original $50$ event concept labels to generate the next event in the sequence (full text of the prompt is provided in the supplementary material). We append the model output to another ICL prompt with \textit{conjunctive} event examples (i.e. questions of the form ``What happens after \texttt{X, Y,} and \texttt{Z}''). We repeat this process until we reach a pre-defined maximum sequence length (in this case, $10$) or until the model fails to generate an in-vocabulary response in $k$ maximum attempts (in this case, $k=3$). In this process, we test the following two ablations -- 

\paragraph{Number of exemplars}
We varied the number of in-context exemplars between 1-12. This is in addition to the zero-shot experiments described in the main paper. We evaluated recall (i.e. proportion of references captured by the resulting output sequences) for all generated outputs in an automated way through matching lexical alignment with Wikidata references. We observed none to marginal  improvements by varying the number of exemplars beyond 3 in the initial trigger prompt, and beyond 5 in the second iterative prompt. 


\paragraph{Selection of specific exemplars}

Selecting in-context examples is an incredibly noisy process \cite{zhang-etal-2022-active}. We started with a static set of randomly selected examples, however owing to Wikidata's inherent label imbalance (political and economic events dominate newsworthy concepts), this led to erratic results, i.e. high recall on similar concepts but not otherwise. 

We then tested a dynamic selection method where every instance of the prompt would contain examples similar to \texttt{target$\_$label}. To achieve this, we used BERT embeddings \cite{devlin-etal-2019-bert} to retrieve (using cosine distance) examples from the reference set. This approach however led to a lower macro-recall and, upon manual inspection of outputs, we observed a high degree of redundancy, i.e. generated outputs were copied from the ICL exemplars. 

To solve for these issues, we reverted to the static prompt examples but manually selected a mix of event topics to be included in the prompt. Our current selection of the prompt yields higher macro-recall than both of the aforementioned techniques tested. While we believe there may be better techniques to select ideal candidates  for ICL exemplars \cite{an-etal-2023-context,agrawal-etal-2023-context} and understand how their compositionality affects specific outputs, we consider such an analysis to be beyond the scope of our work.


We repeat this sequence generation procedure on all $202$ event concepts, generating $2,276$ event sequences with an average length of $5.7$ 
labels per sequence.

\section{Assessing the Quality of LLM-Generated Event Sequences}

Open-ended text generation in a task like event sequencing, with extremely sparse reference data, poses unique challenges to the evaluation of model outputs \cite{wadhwa-etal-2023-revisiting}. The traditional scheme to evaluate discrete model outputs has been to calculate precision and recall for the generated outputs against a predefined reference test set. However, under open-world settings and particularly when a sparse KG like Wikidata is used as reference data, a missing causal relation between two event classes in a sequence may very well be a valid relation. Therefore, it is not possible to automatically derive an accurate measure of precision and recall purely using Wikidata as reference data. We take a multi-pronged evaluation approach to assess the quality and usefulness of the generated event sequences across multiple tasks. This section presents our approach in evaluating the quality of generated event sequences, along with the results of this evaluation. An evaluation of the usefulness of the generated event sequences is presented in Section~\ref{sec:event_sequence_analysis}.

\subsection{Human Evaluation of Cause-Effect Prediction Accuracy of LLMs}
To quantify how well model outputs correlate with human assessments, we first conduct a small-scale human evaluation on a different, independent event-commonsense reasoning dataset: ATOMIC \cite{Hwang2021COMETATOMIC2O}. ATOMIC consists of event-centered pairs of instances of the form \texttt{IsAfter} (\texttt{Y} comes after \texttt{X}) and \texttt{Causes} (\texttt{X} causes \texttt{Y}). We use these instances as input prompts to the model and then ask human annotators to evaluate model outputs. Specifically, we show human annotators anonymized model outputs and true references and elicit their preferences given a trigger event.

To generate outputs, we follow the same strategy as above for a set of $200$ randomly selected event-centered input instances ($100$ each of the type `$X$ \texttt{Causes} $Y$' \textbf{and} `$Y$ \texttt{IsAfter} $X$')\footnote{Complete details about ATOMIC are available in Table 1 in \cite{Hwang2021COMETATOMIC2O}.}. We then present the model output and the true reference from ATOMIC to three human annotators.
For example, a typical instance presented to human evaluators was of the form:

\begin{flushleft}
\texttt{\textbf{Input Instance: }PersonX drops out of high school}

\texttt{\textbf{Response 1: }PersonX gets a job}

\texttt{\textbf{Response 2: }PersonX turns PersonX's life around}

\texttt{\textbf{Type: }IsAfter}
\end{flushleft}

One of the \textit{responses} above is the LLM output, while the other is the true reference. The human evaluators are then asked to answer the following questions-- 
\begin{itemize}
    \item Are both responses \textit{functionally} similar?
    \item Which response do you prefer?
    \item Which response, if any, is completely irrelevant?
\end{itemize}
We find that in an overwhelming majority of the cases, the models generate output that is semantically equivalent to the reference (even though there is no direct lexical alignment), or output that the humans prefer over the true reference. Based on the responses from human evaluators\footnote{We observe strong inter-rater agreement with a Fleiss kappa, $\kappa=0.81$; conflicting response labels were aggregated through majority vote.}, we observe that humans found $65.82\%$ of response pairs \textit{functionally} equivalent. That is, even though not lexically aligned, they meant the same thing. In $27.64\%$ of the instances the humans 
preferred the model-generated event instance over the true reference. In only $6.55\%$ of instances did the humans prefer the true reference over the model-generated output.\footnote{Additional details on these experiments are available in the supplementary material (Appendix).} These results reinforce our underlying assertion that LLMs are capable of event-centered reasoning, and therefore could produce high-quality event sequence collections.

\subsection{Evaluation of Recall} 
Despite the sparsity of causal relations in Wikidata, one can still reasonably estimate recall through pairwise comparisons of generated event classes to the existing causal relations in Wikidata. We build a reference set of causal relations in Wikidata by curating a list of all the pairs of event concepts that have any of the several causal relations\footnote{\url{www.wikidata.org/wiki/Wikidata:List_of_properties/causality}} in Wikidata in any direction, including {\tt has\_cause (P828)}, {\tt has\_immediate\_cause (P1478)}, {\tt has\_contributing\_factor (P1479)}, and  {\tt has\_effect(P1542)}. Our results show that this is an effective mechanism of comparing recall across methods, as larger models that are expected to have better accuracy yield higher recall scores.

\subsection{Evaluation of Precision} 
To overcome  sparsity in reference data, prior work has generally relied on human evaluations \cite{chiang-lee-2023-large} 
for estimating precision, which entails looking at pairs of events and asking 
annotators whether or not the events in question have a causal relationship. Such a process for potentially thousands of event pairs (like in our case) can be very cost prohibitive. Furthermore, recent research \cite{zhao-etal-2022-lmturk,he2023annollm,zheng2023judging} indicates that pre-trained language models themselves might outperform lay human annotators such as those found on Amazon Mechanical Turk. For instance, \cite{he2023annollm} demonstrated that labeling data under a few-shot chain-of-thought prompt (``explain-then-annotate'' setting) surpasses crowdworker annotations on relevance assessments. Following these results, we use LLMs for evaluating precision and for selecting the most optimal model. We propose evaluating precision for model selection as a binary classification task. Given two events $e_1, e_2$, an \textit{evaluator} model must evaluate whether $e_1$ reasonably leads to $e_2$ under a \texttt{has\_cause} or \texttt{has\_effect} relationship. To do this, we start with a large instruction-tuned model (in our case, Flan-T5-XXL ($11$B) \cite{chung2022scaling}) and adopt instructional in-context few shot prompting to classify whether or not $e_1$ reasonably leads to $e_2$, and for the model to provide a justification for its results. While we hypothesize that such an approach may yield noisy results, a small manual assessment as well as our results comparing different models with different evaluator models indicate that this approach yields reliable results for comparing different models, and a reasonable estimate of the overall precision of the model. Note that the evaluator performs only a simple binary classification task that has a very high accuracy even in smaller models.

\subsection{Settings}
We performed all our model inference related experiments on two NVIDIA V100 GPUs. We used the Huggingface library v$4.26.1$ \cite{wolf-etal-2020-transformers} and publicly available model checkpoints.\footnote{\url{huggingface.co/docs/transformers/model_doc/flan-t5}} Classical pattern mining algorithms (GSP, SPADE) were implemented  in Python through the use of \texttt{spmf}\footnote{pypi.org/project/spmf/} library v$1.4$. For event sequence generation, we use in-context learning under zero shot settings for all LLMs, with \texttt{top-k} sampling ($k=50$, in conjunction with \texttt{top-p}, where $p=0.95$). To identify influencing events from Summary Markov Models (SuMMs), we use implementations from \cite{ijcai2022p0670} and split the dataset (generated from LLMs) into train/dev/test sets ($70\%/15\%/15\%$) to generate results reported in Table \ref{tab:res_nlloss}. BSuMMs and OSuMMs were learned with hyperparameters of $\alpha=0.1$, $\gamma=0.5$ and a look-back ($\kappa$) window of $4$.

\subsection{Results}
Table \ref{tab:llm_metrics} summarizes our results from these experiments. 
While prior work as proven the concept of LLM-as-a-judge~\cite{zheng2023judging} when much larger LLMs like GPT-4 are used, here we use less resource-intensive LLMs not for evaluation of the absolute quality of the generated sequences, but for comparison of the relative performance of models

Since we use our models for dual purposes -- for generating \textit{and} evaluating event sequences, owing to this inherent conflict we find it prudent to independently assess these evaluator models. To ensure  robustness of our results, we use multiple evaluator models and find no significant difference between evaluated precision across different models used as evaluators with the largest model performing marginally better as a precision evaluator.

\begin{table}[t]
\small
\centering
\begin{tabular}{@{}lrrrrr@{}}
\toprule
\multicolumn{1}{r}{} & \multicolumn{3}{c}{\textbf{Precision Evaluator Model (P)}}                       & \multicolumn{1}{c}{\textbf{R}} & \multicolumn{1}{c}{\textbf{F-1}} \\ \cmidrule(l){2-6} 
\multicolumn{1}{r}{} & \textit{Flan-T5-Large} & \multicolumn{1}{c}{\textit{Flan-T5-XL}} & \multicolumn{1}{c}{\textit{Flan-T5-XXL}} & \multicolumn{1}{l}{}  & \multicolumn{1}{l}{}    \\ \cmidrule(lr){2-4}
Flan-T5-Large ($783$M) \cite{chung2022scaling}          & 0.73         & 0.72                          & 0.72                           & 0.47                    & 0.57                      \\
Flan-T5-XL (3B)             & 0.75         & 0.75                          & 0.78                          & 0.49                    & 0.60                      \\
Open-Research LLaMA ($3$B) \cite{touvron2023llama}      & 0.70         & 0.69                          & 0.72                           & 0.45                    & 0.55                      \\
Flan-T5-XXL   ($11$B)          & \textbf{0.79}         & \textbf{0.79}                          & \textbf{0.81}                           & \textbf{0.54}                    & \textbf{0.65}                      \\ \bottomrule
\end{tabular}
\caption{Evaluation of LLM-generated sequences. For the purpose of evaluating recall (R), we count event pairs \texttt{(e$_1$, e$_2$)} if such a pair exists in Wikidata. For evaluating precision (P), we treat correctness of all generated event pairs \texttt{(e$_1$, e$_2$)} as a standard classification task. Best-performing model scores are in bold.}
\label{tab:llm_metrics}
\end{table}

\subsection{Discussion}
We observed in our analysis that some event triggers did not produce any sequences. Out of the $202$ unique input event triggers in our data, $8$ event triggers yielded sequences of length $1$, i.e., no sequences. 
This means that the underlying model could not select a relevant event concept from the provided vocabulary that might follow the given trigger.

\section{Knowledge Distillation Through Event Sequence Analysis}
\label{sec:event_sequence_analysis}

Given a high-quality collection of event sequences generated by utilizing LLMs, we use pattern mining to discover high-support sequence patterns not directly derivable from the knowledge graph, and learn probabilistic models to discover complex event sequence rules.

\subsection{Mining New Patterns}
In order to derive new, unseen relationships (\texttt{has\_cause} or \texttt{has\_effect)} between the extracted event classes, we use the generated event sequence collection followed by classical frequent itemset mining algorithms like GSP \cite{10.1007/BFb0014140} and SPADE \cite{Zaki2004SPADEAE} to derive new high support patterns. 
This aspect of the overall workflow is illustrated in the middle section of Figure \ref{fig:main}, where high-support patterns are mined from event sequences.

The classical pattern mining algorithms we use to mine for new event patterns are both Apriori-based approaches. Under both methods, given two sequences of the same event concept class
\begin{equation}
    \alpha=<a_1, a_2,..,a_n> \text{ and }  \beta=<b_1, b_2,..,b_n>
\end{equation}
$\alpha$ is a subsequence of $\beta$ denoted as $\alpha \subseteq \beta$ iff there exists a set of values $1\leq j_1<j_2<...<j_n \leq m$ such that $a_1 \subseteq b_{j1}, a_2 \subseteq b_{j2}, ..., a_n \subseteq b_{jn},$; then $\beta$ is a supersequence of $\alpha$. Given multiple such sequences and a support threshold, the task here is to find a set of frequent event subsequences. GSP is depicted in Algorithm~\ref{alg:algorithm}). SPADE uses a "vertical database format", which stores sequences as lists of itemsets associated with their IDs (referred to as ``id-lists"), which allows faster support counting. The algorithm organizes the search space into equivalence classes, and avoids multiple scans of the sequences.


\begin{algorithm}
\caption{Generalized Sequential Pattern (GSP) Mining Algorithm}
\begin{algorithmic}[1]
\State \textbf{Input:} Database of sequences $D$, minimum support threshold $\text{min\_sup}$
\State \textbf{Output:} Set of all sequential patterns $SP$
\State $SP \gets \emptyset$
\State $C_1 \gets $ set of all individual items in $D$ that meet $\text{min\_sup}$
\State $L_1 \gets $ filter candidates in $C_1$ by $\text{min\_sup}$
\State $k \gets 2$
\While{$L_{k-1} \neq \emptyset$}
    \State $C_k \gets $ generate\_candidates($L_{k-1}$)
    \For{each sequence $s \in D$}
        \For{each candidate $c \in C_k$}
            \If{$c$ is contained in $s$}
                \State increment support count of $c$
            \EndIf
        \EndFor
    \EndFor
    \State $L_k \gets $ filter candidates in $C_k$ by $\text{min\_sup}$
    \State $SP \gets SP \cup L_k$
    \State $k \gets k + 1$
\EndWhile
\State \textbf{return} $SP$
\end{algorithmic}
\label{alg:algorithm}
\end{algorithm}





\subsubsection{Examples of Mined Patterns}
Following are example outputs from application of GSP on LLM outputs. 

\paragraph{Mined Pattern Example 1}
\begin{flushleft}
\texttt{\textbf{Pattern:} Civil Disorder \textit{(Q686984)} $\rightarrow$ Democratization \textit{(Q1064441)} $\rightarrow$ Energy Crises \textit{(Q8413663)}} 
\texttt{\textbf{Support:} 5}
\end{flushleft}

\noindent The pattern above occurs with a support value of $5$ i.e. at least supported by $5$ super-sequences. Empirically, we can find support for such a pattern in history.\footnote{\url{wikipedia.org/wiki/South_African_energy_crisis}} In 1994, the country of South Africa democratized post civil disorder. This led to an increased energy demand over the following decade, eventually culminating in a full blown energy crises starting 2003. 

\paragraph{Mined Pattern Example 2}
\begin{flushleft}
\texttt{\textbf{Pattern:} Famine \textit{(Q168247)} $\rightarrow$ Refugee Crises \textit{(Q20898283)} $\rightarrow$ Post Traumatic Stress Disorder\textit{(Q202387)}} 

\texttt{\textbf{Support:} 5}
\end{flushleft}

\noindent The pattern above again occurs with a support value of $5$ i.e. at least supported by $5$ super-sequences. Similar to the previous example, evidence supporting such an event tranisiton can be found in real life.\footnote{\url{wikipedia.org/wiki/Great_Famine_(Ireland)}} During the Great Irish Famine people were forced to relocate and flee Ireland causing a refugee crises. A great number of these individuals suffered from mental health crises (e.g. PTSD) due to the events directly associated with the famine.

\subsection{Identifying Influencing Sets through Summary Markov Models}
Learning about potential \textit{influencing events} that lead to a given event type in a large set of event sequences is an important aspect of analyzing multivariate event sequences, i.e. events \textit{without time stamps}, like the ones generated by our models. Classic $k_{th}$ order Markov chains capture these dynamics by modeling the probability of observing a particular event type given the preceding $k$ events in-sequence. 
The recent family of summary Markov models (SuMMs) \cite{ijcai2022p0670} generalize other well known Markov models for event sequences by leveraging a function 
that summarizes historical event occurrences, and identifying the subset of event types that affect the probability of occurrence of event types of interest; this forms the influencing set.


We use the LLM-generated sequences to learn two types of SuMMs: binary SuMMs (BSuMMs) and ordinal SuMMs (OSuMMs). In BSuMMs, it is only the presence or absence of an event in the relevant history that has an effect on the occurrence of other events, while in OSuMMs the order of the events is also taken into account. We refer the reader to Sections 3.3 and 3.4 in \cite{ijcai2022p0670} for complete formal definitions and methods for learning SuMMs over event sequence collections. Briefly, given a subset of event labels of interest $\mathbf{X} \in \mathcal{L}$ and a set of parameters 
$\Theta_X = \{ \theta_{x|h} \}$, where $\theta_{x|h}$ is the probability of a given event label $x \in \mathbf{X}$ occurring at any position in the sequence given the historical event occurrences $h$, \textit{influencing} and \textit{non-influencing} event sets can be formally defined as label sets \textbf{U} = $\mathcal{L}\setminus$ \textbf{U} are influencing sets for event labels $\mathbf{X}$ such that they minimally determine the probability of observing any particular label of interest $x_i$ for a given position $i$ in the sequence.

\subsubsection{Evaluation: SuMMs vs LSTMs}
Following the evaluation strategy implemented in \cite{ijcai2022p0670}, we focus on individual labels of interest and conduct an evaluation around probabilistic prediction. Consequently, we select negative log loss as the evaluation metric. Table \ref{tab:res_nlloss} summarizes our results on the test set for both BSuMMs and OSuMMs, along with a simple LSTM baseline. We observe that models trained on event sequences generated from larger models (e.g. Flan-T5-XXL) fare better than the ones generated from their smaller counterparts (e.g. Flan-T5-Large). We treat this observation as a proxy for generated event sequence quality. That is, better quality sequences lead to better predictive models. 

\begin{table}[th]
\small
\centering
\begin{tabular}{@{}lrrr@{}}
\toprule
($\downarrow$) \textbf{Data Generator Model}   & \multicolumn{1}{c}{\textbf{BSuMMs}} & \multicolumn{1}{c}{\textbf{OSuMMs}} & \multicolumn{1}{c}{\textbf{LSTM}} \\ \midrule
Flan-T5-Large (783M)  \cite{chung2022scaling}          & -63.49            & -84.29                     & -109.28                                    \\
Flan-T5-XL (3B)            & -63.20                     & -92.65                     & -121.23                            \\
Open-Research LLaMA (3B) \cite{touvron2023llama} & -110.57                    & -101.29                 & 
-190.68 \\
Flan-T5-XXL       (11B)       & \textbf{-57.99}            & \textbf{-78.64}                  & \textbf{-108.89}                                    \\ \bottomrule
\end{tabular}
\caption{Negative log likelihood 
(\textit{lower magnitude is better}) 
averaged over interest labels from LLM-generated (Flan-T5-XXL) event sequences. Lookback window ($k$) for LSTM was fixed to $5$. Best-performing data generator model scores are in bold.}
\label{tab:res_nlloss}
\end{table}

\subsubsection{Qualitative Assessment } 
Figure~\ref{fig:main} shows examples of learned influencing sets using BSuMMs for {\tt refugee crisis} and {\tt mass migration} events. Two events {\tt mass migration} and {\tt famine} are identified as influencing events for {\tt refugee crisis}. The model indicates that the occurrence of both {\tt mass migration} and {\tt famine}  together has a $0.91$ probability of resulting in {\tt refugee crisis} as a part of a sequence of events. On the other hand the occurrence of {\tt mass migration} in the absence of {\tt famine} has a $0.31$ probability of resulting in  {\tt refugee crisis}. BSuMMs and OSuMMs deploy a greedy score-based forward and backward search strategy to efficiently discover the minimal influencing sets. Overall, the discovery of influencing sets from LLM-generated data provides a mechanism of distilling complex symbolic knowledge from the output of LLMs.

We provide two more examples below of influencing sets derived from the application of SuMMs. For each example, we show evidence supporting the accuracy of the extracted knowledge. Overall, these results show the high quality and usefulness of the distilled event sequence knowledge.

\paragraph{Influencing Set Example 1}
The following example identifies ``Hate Crimes'' as a predecessor to the occurrence of ``Civil Disorder'' related events. 
\begin{flushleft}
\texttt{\textbf{X:} Civil Disorder \textit{(Q686984)}} 

\texttt{\textbf{Parent:} Hate Crime \textit{(Q459409)}} 

\begin{itemize}
    \item $P(\text{\texttt{Civil Disorder}} | \text{\texttt{NO Hate Crime}}) = 0.12$
    \item $P(\text{\texttt{Civil Disorder}} | \text{\texttt{Hate Crime}}) = 0.55$
\end{itemize}
\end{flushleft}

In the example above, we see that the likelihood of a civil disorder is greatly influenced by the occurrence of a hate crime. 

\paragraph{Influencing Set Example 2}
The following example identifies ``Disease Outbreaks'' and ``Lockdowns'' as precursors to the institution of ``Travel Restrictions''. 
\begin{flushleft}
\texttt{\textbf{X:}Travel Restriction  \textit{(Q87745167)}} 

\texttt{\textbf{Parents:} Disease Outbreak \textit{(Q3241045)}, Lockdown \textit{(Q6665312)}} 

\begin{itemize}
    \item $P(\text{\texttt{TR}} | \text{\texttt{NO Outbreak, NO Lockdown}}) = 0.0.0001$
    \item $P(\text{\texttt{TR}} | \text{\texttt{Lockdown, NO Outbreak}}) = 0.0.29$
    \item $P(\text{\texttt{TR}} | \text{\texttt{NO Lockdown, Outbreak}}) = 0.26$
\end{itemize}
\end{flushleft}

Quite intuitively, we see here the likelihood of Travel Restrictions directly correlate with the institution of Disease Outbreaks and Lockdowns (since the latter have been lately associated with the former). 

\subsubsection{Error Analysis}
We observed that in some cases, the identified influencing sets
for some event concepts were not correct or, at times, illogical. Qualitatively, we observe that this occurs with rare event types that do not appear often either in the underlying knowledge graph, or in the LLM-generated set of event sequences. Consequently, SuMMs fail to identify logical influencing sets. For example, for the event type \texttt{United States Presidential Impeachment}, BSuMM model identified  \texttt{unequal treaty} as the influencing event. 

\section{Conclusions and Future Work}
In this paper, we presented methods of distilling event sequence knowledge from large language models. We first presented an approach of generating high-quality event sequences through knowledge-guided generation using LLMs. 
We then conducted a three-pronged quantitative and qualitative evaluation of our results. First, we evaluated LLM-generated event sequences in absolute terms through manual evaluation and automated evaluation using standard accuracy metrics. Second, we looked at some qualitative examples of \textit{newly} identified high support sequences not already present in our base knowledge graph i.e. Wikidata. Finally, we gauge how probabilistic models like SuMMs fare when trained on LLM-generated data. We found that through our carefully-designed elicitation procedure, LLMs are capable of producing high-quality event sequences that can be used for the downstream tasks of pattern mining and the construction of probabilistic event models such as SuMMs. 

This work demonstrates the utility of large language models for extracting event-related information and helps researchers better navigate the complex interaction between event-related entities. While we have demonstrated some use-cases for the resulting dataset of event sequences -- mining logical rules/patterns and extracting influencing event types -- the resulting datasets in this work and those from other domains could themselves be a useful resource for researchers, as one may leverage them suitably to inform or justify decision-making. We make our code and datasets publicly available for future research. Here, we outline a number of directions for future work:

\begin{itemize}
    \item In this paper, we used Wikidata as our base source of knowledge. The event concepts found on Wikidata often appear in the pre-training datasets of large language models. We did not consider more complex, domain-specific datasets of non-timestamped events (e.g. Healthcare, Finance). Applying LLMs to generate domain-specific events may require considerable amounts of data to fine-tune these models, and collecting such supervision to train these models at scale may be prohibitively expensive.
    \item A key element of our evaluation strategy (precision) involves the use of the same language models that were used to generate the sequences being evaluated in the first place. Through qualitative examples and past research on using LLMs as annotators, we observe that such a strategy yields a noisy but useful signal for evaluating model performance. However, an ideal and accurate evaluation of model outputs should involve the use of human annotators and a validation of the use of larger LLMs as a part of LLMs-as-a-judge~\cite{zheng2023judging} evaluation strategy. Furthermore, our use of smaller models was justified in part due to the simplicity of the precision evaluation task, which can be posed as a binary classification task. It will be interesting to evaluate the correctness of post-hoc explanations generated by models for their corresponding output during precision evaluation, as such explanations are key to the usefulness of the distilled knowledge, particularly in critical decision support applications.
    \item In the context of our domain of interest in this paper (news event analysis), an exciting avenue for future research is to study the effect of utilizing structured knowledge distilled from LLMs in a pipeline for future event prediction, and a comparison with the performance of human forecasters. Recent work~\cite{halawi2024approaching} claims that LLMs, without symbolic knowledge elicitation, come close and in some cases surpass the ability of a ``crowd aggregate of competitive forecasters" in making accurate forecasts in the form of answering forecast questions on online platforms such as the Good Judgment Open (\url{(https://www.gjopen.com)}). Two research questions arise: 1) Could a neurosymbolic approach utilizing distilled symbolic knowledge complement or outperform a purely LLM-based solution? 2) Would distilled high-quality structured knowledge provide better explainability and therefore be a more reliable tool as a part of a human-AI collaboration mechanism for event forecasting?
\end{itemize}

\paragraph*{Supplemental Material Statement:} Our supplementary material (included as a zip file in our submission) includes code and the prompts used for event sequence generation and benchmarking, as well as the base KG, and sample output files. {\tt README.md} has all the details. {\tt Appendix.pdf} contains our prompts and human evaluation details.

\bibliographystyle{splncs04}
\bibliography{references,custom, anthology}
\clearpage

\section*{Appendix}
This section constitutes technical appendix (i.e. supplementary material) for the submission titled ``Distilling Event Sequences from Large Language Models''. 

\section{Event Sequence Prompts}
\label{sec_apx:event_seq_prompts}
We now describe the specific prompt-types used (including pseudo-code for their use) to generate event sequences given individual Wikidata event concepts. We start a Wikidata event concept label and feed it into the following prompt with 3 ICL exemplars. The prompt also includes a label space of other Wikidata concepts from which we i nstruct the model to choose a concept. In case of an out-of-domain generated output ($\approx 15\%$ of total outputs), we discard those outputs. \linebreak

\begin{lstlisting}[language=Python, caption=Initial Trigger Prompt]
def build_prompt1(vocab, target_label):
    prompt = "Use the following vocabulary to respond to the questions: " + \
           f"{' '.join(label_space)}\n" + \
           f"Question: what usually happens after earthquake?\n" +\
           f"Answer: tsunami\n" + \
           
           f"Question: what usually happens after economic crises?\n" +\
           f"Answer: unemployment\n" + \
           
           f"Question: what usually happens after bomb attack?\n" +\
           f"Answer: injury\n" + \
           
           f"Question: what usually happens after {target_label}?\n" +\
           f"Answer:"
           
    return prompt
\end{lstlisting}

Following the output from the above prompt trigger, we feed the in-domain outputs to the following ICL iterative prompt with \textit{conjunctive} event exemplars (i.e. \texttt{X and Y}). We then successively use this prompt by feeding model generated outputs as new event concepts. Restricting the generated vocabulary to existing Wikidata concepts allows for a reasonable recall evaluation even from a sparse reference set. We illustrate this approach in Figure \ref{fig:prompt_examples} in the main paper. \linebreak
\begin{lstlisting}[language=Python, caption=Iterative ICL Prompt with Conjunctive Examples]
def build_prompt2(vocab, target_labels):

    prompt = "Use the following vocabulary to respond to the questions: " + \
           f"{' '.join(label_space)}\n" + \
           f"Question: what usually happens after earthquake?\n" +\
           f"Answer: tsunami\n" + \
           
           f"Question: what usually happens after earthquake and tsunami?\n" +\
           f"Answer: nuclear disaster\n" + \
           
           f"Question: what usually happens after economic crises and wage decline and unemployment?\n" +\
           f"Answer: legislation\n" + \
           
           f"Question: what usually happens after military conflict?\n" +\
           f"Answer: war\n" + \
           
           f"Question: what usually happens after military conflict and war?\n" +\
           f"Answer: peace treaty\n" + \
           
           f"Question: what usually happens after {' and '.join(target_labels)}?\n" +\
           f"Answer:"

    return prompt
\end{lstlisting}

\begin{figure*}[t]
\centering
  \includegraphics[width=\textwidth]{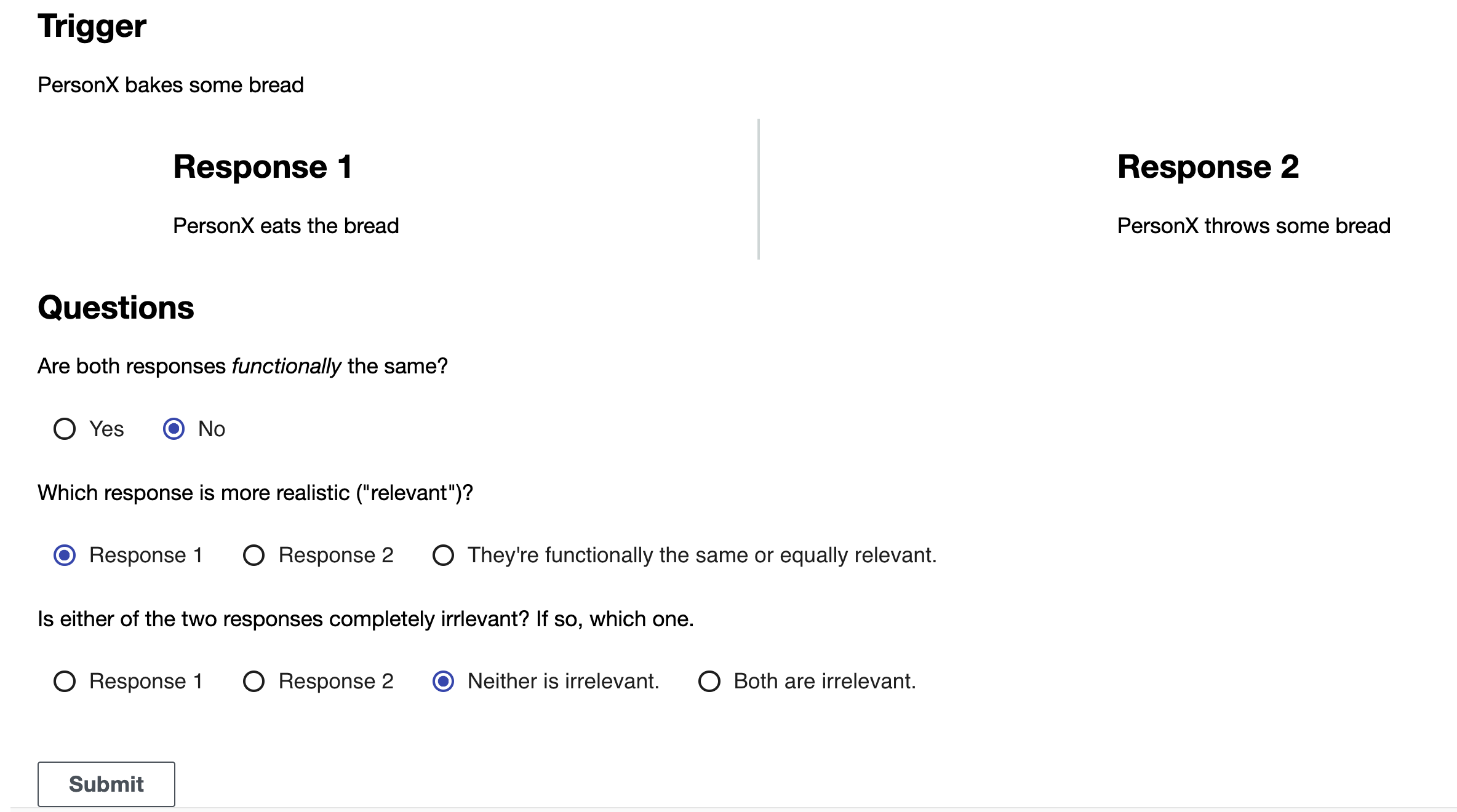}
  \caption{
  Annotation interface for classifying model generated response vs reference (anonymized) for the ATOMIC event descriptions.  
  }
  \label{fig:amt}
\end{figure*}

\section{Evaluator Models (Example Outputs and Prompts)}
For a model generated event sequence $\alpha=a_1, a_2,..,a_n$, we consider all possible pairs of events $(a_i, a_{i+1})$ as \texttt{(trigger, consequence)} pairs where $a_0$ is the initial trigger sequence and $i\in{1,..,n}$ are all events generated by the model that follow $a_0$. Then we use the following prompt to evaluate correctness of all such possible pairs and use this as proxy for a true precision evaluation of the generated sequences. \linebreak
\begin{lstlisting}[language=Python, caption=Prompt used to evaluate correctness of a given event pair]
def build_precision_eval(trigger, consequence):
    prompt = "Respond to the questions below with a (YES/NO) with a historical example:" + \
            f"Question: Can economic crises cause a landslide?\n" +\
            f"Answer: NO. There is no historical example of an economic crisis causing a landslide, which is natural disaster.\n" + \
            f"Question: Can earthquake cause a tsunami?\n" +\
            f"Answer: YES. In 2011, Japan experienced an earthquake in tohoku that caused a tsunami. \n" + \
            f"Question: Can mass shooting cause a condensation cloud?\n" +\
            f"Answer: NO. A condensation cloud is a weather phenomenon, not a mass shooting.\n" + \
            f"Question: Can accident cause a stock market crash?\n" +\
            f"Answer: NO. The stock market crash of 1929 was caused by a series of events, not an accident.\n" + \
            f"Question: Can disease outbreak cause a inventory shrinkage?\n" +\
            f"Answer: YES. The bubonic plague outbreak in Europe in 1348 caused a massive inventory shrinkage.\n" + \
            f"Question: Can fraud cause a travel ban?\n" +\
            f"Answer: YES. Travel bans are a form of punishment for immigration fraud.\n" + \
            f"Question: Can {trigger} cause a {consequence}?\n" + "Answer: "
    return prompt
\end{lstlisting}

As mentioned in the limitations section, a key shortcoming of our approach here is that we do not evaluate the correctness of the post-hoc explanations generated by the model for it's classification label. We leave that analysis for future work.

\section{ATOMIC Human Evaluations}
Amazon Mechanical Turk (AMT) is a platform for non-expert works to perform microtasks, in our case -- human annotations. Three authors with graduate degrees in computer science carried out these human evaluations. Figure \ref{fig:amt} shows the interface provided to these human annotators. As described in the main paper, human evaluators were asked to answer the following questions-- 
\begin{itemize}
    \item Are both responses \textit{functionally} similar?
    \item Which response do you prefer?
    \item Which response, if any, is completely irrelevant?
\end{itemize}

\paragraph{Functionally Similar Responses} For responses to be functionally similar, they \textit{must} convey the same meaning even if there is major lexical mismatch between the actual tokens. 

\paragraph{Human Preferences} The preferences elicited here are based on a humans degree of reasonableness given an event trigger. We observed a high Fleiss $\kappa=0.81$ indiciating a high degree of agreement between the annotators. 

\paragraph{Irrelevant Responses} Here, again the annotators were asked to exercise judgment in what they may find a \textit{completely unreasonable} response to a given event trigger. 

We release the results from these human annotations for a comprehensive analysis and any additional findings that readers may infer.

\end{document}